\pdfoutput=1

\documentclass[11pt]{article}

\usepackage[final]{acl}

\usepackage{times}
\usepackage{latexsym}

\usepackage[T1]{fontenc}

\usepackage[utf8]{inputenc}

\usepackage{microtype}

\usepackage{inconsolata}

\usepackage{graphicx}

\usepackage{paralist}
\usepackage{inconsolata}
\usepackage{tabularx}
\usepackage{float}
\usepackage{caption}
\usepackage{graphicx}
\usepackage{multirow}
\usepackage{booktabs}
\usepackage{setspace}
\usepackage{booktabs}
\usepackage{paralist}
\usepackage{graphicx} 
\usepackage{amsmath}
\usepackage{colortbl}
\usepackage{booktabs}
\usepackage{multirow}
\usepackage{adjustbox}
\usepackage{array}
\usepackage{siunitx}
\usepackage{xcolor}  
\usepackage{listings}
\usepackage{amsfonts}

\usepackage{arydshln}
\usepackage{caption}     
\usepackage{longtable}   
\usepackage{lscape}

\usepackage[table,xcdraw]{xcolor}

\title{Speaking at the Right Level: Literacy-Controlled Counterspeech Generation with RAG-RL}

\author{First Author \\
  Affiliation / Address line 1 \\
  Affiliation / Address line 2 \\
  Affiliation / Address line 3 \\
  \texttt{email@domain} \\\And
  Second Author \\
  Affiliation / Address line 1 \\
  Affiliation / Address line 2 \\
  Affiliation / Address line 3 \\
  \texttt{email@domain} \\}

\author{
  \textbf{Xiaoying Song\textsuperscript{1}}
  \textbf{Anirban Saha Anik\textsuperscript{1}}
  \textbf{Dibakar Barua\textsuperscript{1}} \\
   \textbf{Pengcheng Luo\textsuperscript{2}}
 \textbf{Junhua Ding\textsuperscript{1}}
  \textbf{Lingzi Hong\textsuperscript{1}}
\\
  \textsuperscript{1} University of North Texas
\\
\textsuperscript{2} Peking University
\\
  \small{
  \{xiaoyingsong, anirbansahaanik, dibakarbarua\}@my.unt.edu} \\ 
  \small{
  luopc@pku.edu.cn,
\{junhua.ding, lingzi.hong\}@unt.edu  
  }
}

\begin{document}
\maketitle
\begin{abstract}
Health misinformation spreading online poses a significant threat to public health.
Researchers have explored methods for automatically generating counterspeech to health misinformation as a mitigation strategy.
Existing approaches often produce uniform responses, ignoring that the health literacy level of the audience could affect the accessibility and effectiveness of counterspeech. 
We propose a \textbf{Controlled-Literacy} framework using retrieval-augmented generation (RAG) with reinforcement learning (RL) to generate tailored counterspeech adapted to different health literacy levels. 
In particular, we retrieve knowledge aligned with specific health literacy levels, enabling accessible and factual information to support generation.
We design a reward function incorporating subjective user preferences and objective readability-based rewards to optimize counterspeech to the target health literacy level. 
Experiment results show that Controlled-Literacy outperforms baselines by generating more accessible and user-preferred counterspeech.
This research contributes to more equitable and impactful public health communication by improving the accessibility and comprehension of counterspeech to health misinformation. 
\end{abstract}

\section{Introduction}
Health misinformation online poses a serious risk to public health as it can mislead people, cause bad health choices, and reduce trust in doctors and medical advice~\cite{do2022infodemics}. A user on Reddit once claimed that ``I won't take a mammogram because it is the squishing that causes cancer.'' Such health misinformation not only misleads individuals into avoiding appropriate medical actions but also increases public confusion~\cite{fridman2023health}. Therefore, it is essential to curb the spread of health misinformation.

\begin{figure}
    \centering
    \includegraphics[width=1\linewidth]{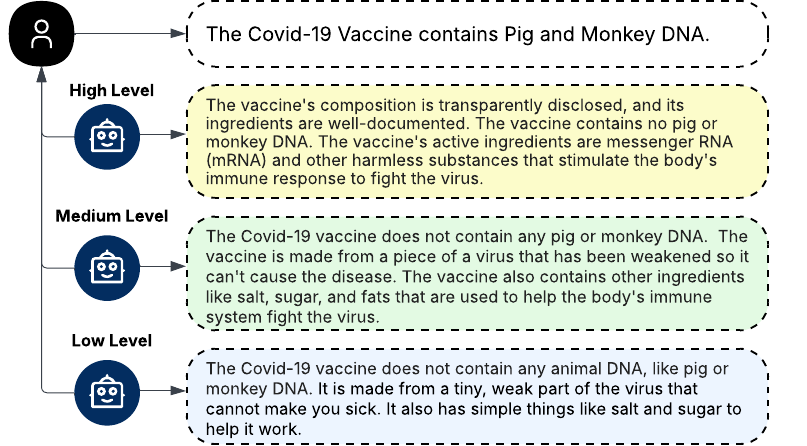}
   \caption{An example of health misinformation, paired with three counterspeech responses tailored to low, medium, and high health literacy levels.}
    \label{fig:Intro_example}
\end{figure}

Social media users actively participate in combating health misinformation by posting counterspeech~\cite{kuo2025predicting,anik2025multi, micallef2020role}. However, user-generated responses often lack credible evidence, decreasing the effectiveness of counterspeech~\cite{yue2024evidence}. 
While experts and fact-checkers can provide factual support in debunking health misinformation, it becomes less efficient when confronted with an increasing volume of health misinformation~\cite{yue2024retrieval}. Generating supportive counterspeech at scale remains a challenge.

Recent studies have explored the use of Large Language Models (LLMs) to generate counterspeech in response to health misinformation~\cite{yue2024evidence,yue2024retrieval,wang2024f2rl,he2023reinforcement}. However, these methods often overlook a critical issue: Is the generated counterspeech accessible and persuasive to users with different health literacy levels? 
Health literacy refers to the capability to understand, judge, and use health information to make good choices and stay healthy~\cite{who2024healthliteracy}. 
The health literacy level of the audience can significantly affect the effectiveness of counterspeech~\cite{liu2020meaning}. 
Users with lower health literacy often struggle to comprehend counterspeech grounded in scientific research~\cite{shahid2022impact,liu2020meaning} and understand information that contains complex or technical language~\cite{cdc2024healthliteracy,chen2018health}. 
In contrast, users with a higher literacy level may lose interest or perceive the content as oversimplified if it is tailored to a low comprehension level~\cite{august2024know,martinez2022optimizing}.
For example, simplistic explanations such as \textit{``It is made from a tiny, weak part of the virus that cannot make you sick ...''} in Figure \ref{fig:Intro_example} (low-level counterspeech) may be inappropriate for users with advanced health literacy. 
Such mismatches in communication can critically undermine the effectiveness of counterspeech. 

In this work, we aim to generate counterspeech for users with diverse health literacy levels. 
To address this question, we propose a controllable RAG framework, the \textbf{Controlled-Literacy}. 
The approach uniquely retrieve knowledge adapting to diverse health literacy levels, considering that the complexity of retrieved knowledge directly influences the style and clarity of the generated responses~\cite{ke2025retrieval}. 
We then integrate Reinforcement Learning (RL) to further optimize the generation, ensuring that the outputs align with the target health literacy level.

Our Controlled-Literacy framework is capable of producing counterspeech that is more accessible, user-preferred, polite, and also factually accurate when addressing health misinformation. We summarize our contribution as follows:
(1) We introduce the novel insight that effective counterspeech should be aligned with the health literacy level of its target audience. 
(2) We propose \textbf{Controlled-Literacy}, a framework that integrates RAG and RL for generating accessible counterspeech that accounts for both \textit{objective readability and subjective user preferences.}
(3) A new dataset is curated, \textbf{MisinfoLiteracy}, consisting of health misinformation posts paired with counterspeech responses tailored to users with different health literacy levels.

\section{Related Work}

\subsection{Counterspeech to Misinformation Generation}
Counterspeech has been proven effective in mitigating misinformation~\cite{peng2024rescuing,siebert2023effective}. Previous studies focus on generating counterspeech with desirable attributes. 
For example, \citet{anik2025multi} and \citet{yue2024evidence} combined external scientific sources to generate evidence-based counterspeech in response to misinformation. ~\citet{yue2024retrieval} generated factually accurate counterspeech by synthesizing contrastive arguments derived from fact-checking sources. ~\citet{hong2024outcome} adapted the generation of counterspeech according to preferred conversation outcomes, resulting in more positive conversation engagement.  
~\citet{he2023reinforcement} proposed a reinforcement learning-based framework to generate counterspeech with enhanced politeness, factuality, and refutational strength.
Despite these efforts, existing studies have not yet investigated counterspeech generation tailored to users with different health literacy levels. 

\begin{figure*}
    \centering
    \includegraphics[width=0.9\linewidth]{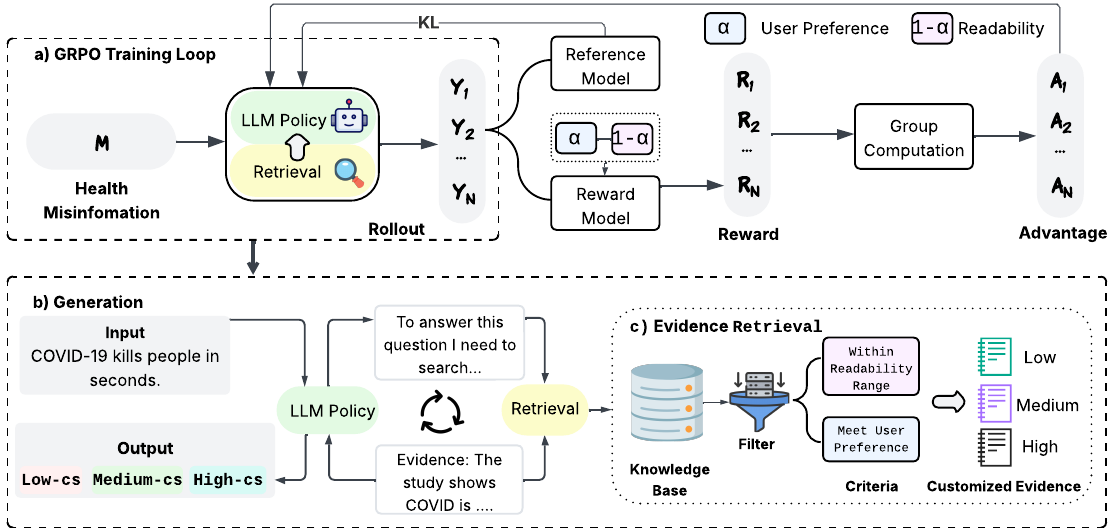}
    \caption{
Overview of our \textit{Controlled-Literacy} counterspeech generation framework tailored to users with different health literacy levels. 
(a) The GRPO training loop integrates evidence retrieval into the LLM policy and optimizes it using a hybrid reward function combining user preference (weight $\alpha$) and readability (weight $1{-}\alpha$). Rewards are aggregated through group computation to compute the advantage signal. 
(b) During inference, the model takes a health misinformation input and retrieves customized evidence to generate counterspeech adapted to low, medium, or high health literacy users. 
(c) The evidence retrieval module selects content from the knowledge base by filtering it according to the target readability range and user preference thresholds, ensuring personalized support for the generation process.
}
    \label{fig:Control_literacy}
\end{figure*}

\subsection{Accessible Language Generation}
Accessible language generation is widely explored in multiple domains, such as health care~\cite{yao2024readme, rahman2024health,luo2022readability,phatak2022medical}, education~\cite{wang2024science,malik2024tarzan, rooein2023know}, and finance~\cite{wang2023fingpt, kosireddy2024exploring,perez2023novel}. In health care, previous studies primarily focus on simplifying or summarizing complex medical terminology. ~\citet{yao2024readme} introduced a dataset with medical terms paired with lay definitions, aiming to enhance the accessibility of medical information for non-experts. In education, studies have investigated tailoring learning materials to learners at varying proficiency levels. For instance, ~\citet{malik2024tarzan} have experimented with methods to control the language proficiency level of LLM-generated content to suit different types of learners. In finance, the emphasis has been on democratizing access to financial information for a broader audience. 
~\citet{kosireddy2024exploring} used small language models to make financial question-answering more accessible for people with limited resources. 
Existing approaches primarily optimize for objective readability metrics while neglecting subjective factors such as user perceptions.

\section{Methodology}
\subsection{Task Definition} 
We follow the framework proposed by~\citet{nutbeam2000health}, which defines and categorizes health literacy into three hierarchical levels:
\emph{Functional Health Literacy} involves essential reading and writing skills necessary for understanding everyday health information.
\emph{Interactive Health Literacy} encompasses more advanced cognitive skills, enabling individuals to communicate effectively and apply new health information to changing circumstances.
\emph{Critical Health Literacy} includes the ability to critically analyze health information and understand determinants of health. 
Based on the framework, we define three target health literacy levels: \textit{low}, \textit{medium}, and \textit{high}, corresponding to Functional Health Literacy, Interactive Health Literacy, and Critical Health Literacy, respectively.

Given the complex definition of health literacy, measuring health literacy in its entirety remains a methodological challenge, especially at scale and in automated settings.
In our framework, we first adopt the Flesch-Kincaid Reading Ease (FKRE) score as a practical, scalable proxy for estimating the functional dimension of health literacy. FKRE quantifies text difficulty based on syntactic features, which directly influence comprehension. This makes it an approximate tool for distinguishing materials accessible to users with varying levels of health literacy. For example, texts with higher FKRE scores (e.g., 80–100) are generally easier to read and more appropriate for individuals with lower health literacy.
Referring to the previous study~\cite{rooein2023know}, we collapse the FKRE score into three categories, easy (80-100), medium (60-79), and hard (0-59). 

However, FKRE alone does not capture users’ ability to critically interpret or act upon health information, dimensions that fall under interactive and critical health literacy. FKRE also fails to reflect whether users perceive the content as helpful, trustworthy, or respectful.
To address this limitation, our approach augments FKRE with simulated user preference ratings, derived from LLMs conditioned to emulate users at different health literacy levels. These simulated raters assess counterspeech based on perceived accessibility, clarity, and helpfulness, providing a subjective complement to the FKRE score. Together, these signals enable our model to generate content that aligns with the syntactic simplicity suitable for a target group and resonates with the group’s communication expectations and information-processing styles.



The task is formulated in the following: given a piece of health misinformation $m$ and relevant retrieved knowledge $k$, the function $f$ generates a counterspeech response $c$ that is tailored to match the health literacy level $l$ of a target user group.
\[
f: (m, k, l) \mapsto c
\]


\subsection{Controlled-Literacy Pipeline}
\label{Controlled-Literacy Pipeline}

The Controlled-Literacy Pipeline includes three sections (See Figure \ref{fig:Control_literacy}): Knowledge Base Construction, Health Literacy-Adaptive Evidence Retrieval, and Controlled Literacy RL Generation.

\textbf{Knowledge Base Construction} 
Users with different levels of health literacy get information from different sources and prefer content with varying complexity and detail \cite{chen2018health}.
We collect knowledge from a diverse set of reliable sources to ensure the inclusiveness and representativeness of knowledge across user groups. These include federal health agencies such as the Centers for Disease Control and Prevention (CDC) and academic and research databases like the Johns Hopkins Children's Center. 
In addition to diverse sources, we employ FKRE to evaluate the readability of each document, categorized as easy (80–100), medium (60–79), or hard (0–59), to ensure that materials are at different readability levels.



\textbf{Health Literacy-Adaptive Evidence Retrieval} 
We utilize a hybrid retrieval method to get evidence from the knowledge base, which has been proven to perform better than a single method ~\cite{sawarkar2024blended}. 
Our retrieval module integrates two retrieval methods: keyword-based  (\(R_k\) ) and semantic retrieval (\(R_s\)). The hybrid retriever (\(R_h\)) integrates the strengths of two methods: $R_h = R_k \cup R_s$.
 When retrieving the top-\( N \) documents (\( R_h = \{d_1, d_2, \dots, d_N\} \)), these documents are concatenated into a single context:
$C = \text{concat}(d_1, d_2, \dots, d_N)$.The concatenated context \( C \) is then paired with the input query \( q \) to construct the prompt for the LLM to generate responses \(r\). 


After retrieving knowledge, we filter the evidence using both the FKRE score and LLM-simulated user preference ratings. We design a 1–5 Likert-style scale to evaluate their preference for the generated counterspeech, where a rating of 3 or higher indicate that users with a certain health literacy level find the content to be at least acceptable.
We retain evidence that falls within the target FKRE range and receives a preference rating of 3 or higher, leaving out content users find unhelpful or confusing (rated 1 or 2). This ensures the selected content is both syntactically accessible and aligned with users’ communication expectations.




\textbf{Controlled Literacy RL Generation} 
A counterspeech generator is trained to adapt to the target audience’s health literacy level. It balances a readability-based reward with a user preference reward to optimize outputs, enabling the generation to be more accessible to a user group. 
The FKRE score assesses the difficulty of texts. Rather than enforcing strict constraints, we consider a response desirable if its FKRE score falls within the corresponding target range. 
This design allows for greater flexibility during optimization and encourages more natural counterspeech generation~\cite{ibrahim2024comprehensive}. We adopt a double-sigmoid function to optimize outputs into the target range of readability:

\begin{equation}
r_{\text{read}} = \sigma\left(\frac{\text{F} - L}{s}\right) - \sigma\left(\frac{\text{F} - R}{s}\right)
\end{equation}


\noindent
\(F\) is the FKRE score.
\(L\) and \(R\) are the left and right boundaries of the target FKRE range. \( s \) is a scaling factor controlling the transition smoothness.

User preference measures whether the generated counterspeech is accessible and easy for users with the target health literacy level to comprehend and apply. 
Since no publicly available datasets exist for this task and employing human annotators is costly, we leverage LLMs with customized instructions to simulate users from different health literacy levels.

We ultimately incorporate both evaluations to optimize generation, ensuring that the counterspeech objectively fits the target readability level and subjectively aligns with user preferences.
We define the final reward as a weighted combination of two components: the readability reward and the user preference reward. Specifically,
\begin{equation}
r(x, y) = \alpha \cdot r_{\text{read}}(x, y) + (1 - \alpha) \cdot r_{\text{pref}}(x, y)
\end{equation}

where \( \alpha \in [0,1] \) is a hyperparameter that controls the balance between promoting text that fits the target readability range and ensuring subjective accessibility for users across different health literacy levels.  $r(x, y)$ is the composite reward.
$r_{\text{read}}$ is the readability reward based on the FKRE score, 
The preference reward $r_{\text{pref}}$ is derived from LLM-based Likert-style scoring (1–5) simulated for a given health literacy level.

After confirming the reward function, we use Group Relative Policy Optimization  (GRPO) ~\cite{shao2024deepseekmath} to optimize the counterspeech. 
GRPO generates $n$ responses for each prompt and computes their relative advantages based on an aggregated reward signal. The policy is then updated to increase the likelihood of higher-ranked responses while constraining divergence from the reference policy $\pi_\phi$ using KL regularization. The optimization objective is:


\begin{equation}
\resizebox{0.48\textwidth}{!}{$
\mathbb{E}_{(x, y_i) \sim \pi_\theta} \left[ r(x, y_i) - \beta\, \mathrm{KL}\left(\! \pi_\theta(y_i \mid x) \,\|\, \pi_\phi(y_i \mid x) \!\right) \right]
$}
\end{equation}

where $r(x, y)$ is the composite reward function, and $\beta$ controls the regularization strength.

\section{Experiment Results}
\subsection{Datasets}
\textbf{MisinfoLiteracy} We utilize PRAW API \footnote{https://praw.readthedocs.io/} to collect health misinformation from Reddit, focusing on topics related to the coronavirus disease (COVID-19). 
We collect Reddit posts containing health-related keywords (e.g., ``vaccines,'' ``COVID-19,'' ``alternative medicine'').
We obtain 3,872 posts from high-engagement subreddits (e.g., r/health or r/science, etc. See details in Appendix \ref{Appendix:SubredditsList}). 
Then, we employ human annotators to identify and filter health misinformation. We document the details in Appendix \ref{Appendix: detecting health misinformation}. The final dataset contained 440 posts labeled as health misinformation. 

\textbf{MisinfoCorrect}
This public dataset contains 789 misinformation tweets processed by \cite{he2023reinforcement} with pairs of misinformation and corresponding counterspeech responses. 
We use this dataset to fine-tune our LLMs to prevent them from rejecting prompts containing misinformation. Additionally, as the dataset originates from a different platform and exhibits a distinct linguistic style, it serves as a testbed for our cross-generalization experiments.

\textbf{Check-COVID}
The dataset is a public benchmark dataset designed to facilitate the fact-checking of COVID-19-related news claims \cite{wang2023check}. The dataset comprises 1,504 claims, which are either directly taken from news articles or manually written by annotators to represent common misinformation. 
The dataset is also used in our cross-generalization experiments.

\subsection{Baseline}
\textbf{Instructional Prompt} We examine whether the generation could achieve the target without further training. We experiment with several prompt settings and document the best-performing prompts in Appendix \ref{appendix: Prompt Design}.


\textbf{RAG} RAG integrates external knowledge into generation to avoid hallucinations. We build a RAG system with evidence selection to generate counterspeech for different health literacy levels. The experiment tests the capability of RAG with no generation optimization. The prompt is derived from the Instructional Prompt setting. 

\begin{table*}[bth!]
\centering
\small
\renewcommand{\arraystretch}{0.7}
\resizebox{0.9\textwidth}{!}{
\begin{tabular}{llcccc}
\toprule
\textbf{Method} & \textbf{Literacy Level} & \textbf{Politeness} & \textbf{Target Distance (↓)} & \textbf{User Preference} & \textbf{Factual Accuracy}\\
\midrule

\multicolumn{5}{l}{\textbf{LLaMA-8B}} \\

Instructional Prompt 
    & low    & 0.39 (0.23) & 2.16 (3.69) & 0.75 (0.03) &0.86\\
    & medium & 0.49 (0.22) & 5.99 (8.24) & 0.74 (0.06) &0.89\\
    & high   & 0.35 (0.19) & 0.07 (0.74) & 0.75 (0.04) &0.87\\
    & \textbf{Avg.} & \textbf{0.41} (0.21) & \textbf{2.74 (4.22)} & \textbf{0.75} (0.04)  & \textbf{0.87}\\

RAG
    & low    & 0.72 (0.25) & 1.30 (3.23) & 0.71 (0.13) &0.89\\
    & medium & 0.66 (0.22) & 4.45 (6.35) & 0.71 (0.13) &0.88\\
    & high   & 0.41 (0.21) & 0.08 (0.75) & 0.72 (0.09) &0.90\\
    & \textbf{Avg.} & \textbf{0.60} (0.23) & \textbf{1.94 (3.44)} & \textbf{0.71} (0.12) & \textbf{0.89}\\

\textbf{\textit{Controlled-Literacy}}

    & low    & 0.84 (0.15) & 1.21 (2.21) & 0.74 (0.07) &0.89\\
    & medium & 0.69 (0.20) & 1.50 (3.81) & 0.73 (0.10) &0.90\\
    & high   & 0.98 (0.02) & 0.00 (0.00) & 0.75 (0.00) &0.93\\
      & \cellcolor{gray!20}\textbf{Avg.} 
    & \cellcolor{gray!20}\textbf{0.84 (0.15)} 
    & \cellcolor{gray!20}\textbf{0.90 (2.01)} 
    & \cellcolor{gray!20}\textbf{0.74 (0.04)} 
    & \cellcolor{gray!20}\textbf{0.91}\\
\midrule

\multicolumn{5}{l}{\textbf{Qwen-7B}} \\

Instructional Prompt 
    & low    & 0.36 (0.22) & 0.37 (3.54) & 0.74 (0.06) &0.71\\
    & medium & 0.48 (0.23) & 4.25 (6.50) & 0.73 (0.13) &0.88\\
    & high   & 0.41 (0.18) & 0.37 (3.54) & 0.73 (0.12) &0.89\\
        & \textbf{Avg.} & \textbf{0.42 (0.15)} & \textbf{1.66 (4.53)} & \textbf{0.73 (0.09)} &\textbf{0.83}\\

RAG 
    & low    & 0.69 (0.25) & 0.60 (1.97) & 0.69 (0.16) &0.79\\
    & medium & 0.52 (0.18) & 3.07 (5.55) & 0.68 (0.18) &0.92\\
    & high   & 0.45 (0.19) & 0.01 (0.13) & 0.73 (0.07) &0.91\\
        & \textbf{Avg.} & \textbf{0.55 (0.13)} & \textbf{1.23 (2.55)} & \textbf{0.70 (0.13)}  & \textbf{0.87}\\

\textbf{\textit{Controlled-Literacy}} 
    & low    & 0.77 (0.18) & 2.09 (3.38) & 0.73 (0.11) &0.84\\
    & medium & 0.55 (0.19) & 4.08 (5.67) & 0.74 (0.08) &0.94\\
    & high   & 0.86 (0.22) & 0.00 (0.02) & 0.75 (0.04) &0.92\\
     & \cellcolor{gray!20}\textbf{Avg.} & \cellcolor{gray!20}\textbf{0.73 (0.16)} & \cellcolor{gray!20}\textbf{2.06 (3.02)} & \cellcolor{gray!20}\textbf{0.74 (0.08)} &\cellcolor{gray!20}\textbf{0.90}\\

\midrule

\multicolumn{5}{l}{\textbf{LLaMA-1B}} \\

Instructional Prompt
    & low     & 0.65 (0.28) & 8.17 (9.36)  & 0.66 (0.17) &0.50\\
    & medium  & 0.40 (0.25) & 17.30 (23.39) & 0.58 (0.22) &0.58\\
    & high    & 0.70 (0.24) & 0.19 (2.87)  & 0.68 (0.14)  &0.54\\
        & \textbf{Avg.} & \textbf{0.58 (0.16)} & \textbf{8.55 (11.87)} & \textbf{0.64 (0.17)} &\textbf{0.54}\\

RAG 

    & low     & 0.50 (0.31) & 16.90 (14.34) & 0.45 (0.25) &0.63\\
    & medium  & 0.41 (0.21) & 9.64 (12.20)  & 0.36 (0.25) &0.71\\
    & high    & 0.42 (0.22) & 0.51 (4.92)  & 0.42 (0.22) &0.63\\
        & \textbf{Avg.} & \textbf{0.44 (0.05)} & \textbf{9.02 (10.49)} & \textbf{0.34 (0.24)}  &\textbf{0.66}\\

\textbf{\textit{Controlled-Literacy}} 

    & low     & 0.73 (0.17) & 2.08 (3.62)  & 0.68 (0.15) &0.61\\
    & medium  & 0.63 (0.21) & 1.86 (4.10)  & 0.71 (0.12) &0.75\\
    & high    & 0.85 (0.27) & 0.00 (0.00)  & 0.67 (0.13) &0.76\\
        & \cellcolor{gray!20}\textbf{Avg.} & \cellcolor{gray!20}\textbf{0.74 (0.11)} & \cellcolor{gray!20}\textbf{1.31 (2.57)} & \cellcolor{gray!20}\textbf{0.69 (0.13)} & \cellcolor{gray!20}\textbf{0.71}\\

\bottomrule
\end{tabular}}
\caption{Counterspeech generation results on \textbf{MisinfoLiteracy} categorized by health literacy levels: low, medium, and high, along with the overall average. The mean(variance) across samples is reported for Politeness, Target Distance, and User Preference. Factual Accuracy reports the percentage of responses that are factually correct.
User preference presents the evaluation by intended user category. 
Higher mean values for politeness, user preference, and factual accuracy indicate better performance, while lower mean values for target distance indicate better alignment. The best overall performance in each category is highlighted in gray.}
\label{tab:Main_results}
\end{table*}

\subsection{Experiment Setup}
We experiment with various LLMs, including models with similar parameter sizes but different architectures (e.g., LLaMA3.1-8B-Instruct\footnote{Available at: \url{https://huggingface.co/meta-llama/Llama-3.1-8B-Instruct}} vs. Qwen2.5-7B-Instruct\footnote{Available at: \url{https://huggingface.co/Qwen/Qwen2.5-7B-Instruct}}), and models from the same family with varying parameter sizes (e.g., LLaMA3.1-8B-Instruct vs. LLaMA3.2-1B-Instruct\footnote{Available at: \url{https://huggingface.co/meta-llama/Llama-3.2-1B-Instruct}}). This setup allows us to investigate model performance across diverse architectures and parameter scales.

\textbf{Knowledge Base}
We collect information related to COVID-19 from diverse sources to construct the knowledge base, including COVID-19 fact sheets from the CDC, materials from Johns Hopkins Children's Center to help kids understand the pandemic, and a collection of research articles about COVID-19 \cite{wang2020cord}. The full list of sources is documented in Appendix \ref{Appendix: Knowledge Bases}.

\textbf{Evidence Retrieval}
We incorporate both keyword-based and semantic retrieval methods. In both cases, the misinformation statement alone is used as the retrieval query, excluding the full prompt to prevent the introduction of irrelevant noise during the retrieval process. Retrieved results from both methods are merged using an \texttt{AND} operation, and the combined candidates are subsequently ranked to select the \textit{Top-$k$} relevant knowledge chunks for generation. We use LLaMA3.1-8B-Instruct as an example and experiment with multiple \textit{Top-$k$} selections (See Appendix \ref{Appendix: Top-k Comparison}). 




\textbf{Evidence Selection}
We filter for evidence that falls within the target FKRE range: easy (80–100), medium (60–79), or hard (0–59), and has a user preference rating equal to or above 3 (the midpoint of a 5-point scale). This ensures the retrieved knowledge is both appropriately readable and aligned with user preferences, enabling more customized and supportive generation.

\textbf{RL Optimization}
We fine-tune LLMs using GRPO to generate counterspeech tailored to users with low, medium, and high health literacy. The policy backbone is a LoRA-adapted supervised model, initially fine-tuned on the MisinfoCorrect dataset, which pairs health misinformation with corresponding counterspeech to prevent response refusal when prompted with misinformation.
For RL, we use the MisinfoLiteracy dataset, splitting it into 80\% for GRPO training and 20\% for inference evaluation. The reward function combines two components:
(1) a double-sigmoid FKRE-based score that promotes readability within a specific target range (e.g., 80–100 for low literacy), and
(2) a GPT-4o-mini-based user preference score, rated on a 1–5 Likert scale, measuring how well the generated counterspeech aligns with user expectations.
These two signals are weighted equally (0.5 readability, 0.5 user preference) and aggregated to compute the total reward. 
For each prompt, LLMs generate four responses and compute their relative advantages based on an aggregated reward signal. The GRPO training loop then updates the policy to increase the likelihood of higher-reward responses while constraining divergence from a fixed reference policy (initialized from the supervised LoRA-adapted model).
This setup enables the model to learn responses that are both accessible and effective across varying levels of health literacy.
The GRPO training is configured with the following hyperparameters: a per-device batch size of 4, gradient accumulation steps of 1, a learning rate of \(5 \times 10^{-6}\), and training epochs of 3. 
For each prompt, the model generates 4 completions with a maximum token length of 200. The GRPO-specific regularization coefficient \(\beta\) is set to 0.2.



\textbf{Generation} For all generations, we set \texttt{max\_new\_tokens} to 200, \texttt{do\_sample=False}, \texttt{temperature=0.5}, and \texttt{top\_p=0.9}.

\subsection{Evaluation}
We evaluate the generated responses across four key dimensions: target distance, user preference, politeness, and factual accuracy.
In the following, we provide detailed definitions and measurements for each of these evaluation dimensions.

\begin{table}[ht]
\centering
\small
\resizebox{0.47\textwidth}{!}{

\begin{tabular}{lcccc}
\toprule
\textbf{Literacy Level} & \textbf{Category} & \textbf{Tolerant Match} & \textbf{Cohen's Kappa} \\
\midrule
\multirow{2}{*}{Low}  
    & Computing & 0.94 & 0.65 \\
    & Human     & 0.96 & 0.75 \\
\midrule
\multirow{2}{*}{Medium}  
    & Computing & 0.96 & 0.65 \\
    & Human     & 0.92 & 0.69 \\
\midrule
\multirow{2}{*}{High}  
    & Computing & 0.96 & 0.65 \\
    & Human     & 0.88 & 0.63 \\
\bottomrule
\end{tabular}}
\caption{Agreement between human-based and computing-based evaluations and inter-annotator agreement among human evaluators on User Preference.}
\label{tab:human_validation}
\end{table}

\textbf{Target Distance} This metric indicates how close the readability of a generated response is to the target range. We use the FKRE score to measure readability. As outlined in Section \ref{Controlled-Literacy Pipeline}, FKRE scores are categorized into three distinct ranges corresponding to different health literacy levels. The target distance quantifies the deviation between the actual FKRE score of a response and its designated target range. A lower target distance indicates better alignment with the intended readability level. 

 \begin{table}[ht]
\centering
\resizebox{0.47\textwidth}{!}{

\begin{tabular}{cccc}
\toprule
\textbf{Literacy Level} & \textbf{Pearson} & \textbf{Spearman} & \textbf{Kendall's Tau} \\
\midrule
Low level   & 0.68 & 0.65 & 0.59 \\
Medium level & 0.67 & 0.60 & 0.56 \\
High level & 0.61 & 0.73 & 0.70 \\
\bottomrule
\end{tabular}}
\caption{Correlation between human ratings and LLM-generated ratings across literacy levels}
\label{tab:correlation}
\end{table}

\textbf{User Preference} This dimension captures how a response is subjectively perceived by users with different health literacy levels. 
It offers additional insight into how well a counterspeech message connects with its intended audience, for example, how people with low literacy levels view the effectiveness of messages designed for them, as well as those designed for medium or high literacy audiences. 
Due to the high cost of recruiting real users with diverse literacy levels, we employ LLMs as evaluators. These LLMs are guided with tailored instructions to simulate users at specific health literacy levels. We use a 1–5 Likert-style scale to assess user preference, with detailed evaluation prompts documented in the Appendix \ref{Appendix: User Preference Prompt}. 
We further conduct human validation for the user preference evaluation (see details in Appendix \ref{Appendix: Evaluation Validation}). 
Before the evaluation, we administer a brief health literacy survey adapted from \citet{rasmussen2023brief}, consisting of several short questions to assess participants’ health literacy levels.
Based on the survey results, we recruit participants representing low, medium, and high health literacy groups to ensure diversity and alignment with target user profiles.
The results in Table~\ref{tab:human_validation} show that the agreement rate among human annotators exceeds 0.85, with Cohen’s Kappa values above 0.60, indicating substantial agreement. The agreement rate between human judgments and LLM-generated evaluations is above 0.90, and the Cohen’s Kappa exceeds 0.60, indicating a reliable evaluation by the LLM evaluator. In addition, we have conducted a correlation analysis between human ratings and LLM-generated ratings of user preference (see Table \ref{tab:correlation} ). The results suggest that the LLM’s behavior adapts well to different user profiles.
We document the analysis in Appendix \ref{Appendix: Evaluation Validation} Correlation Analysis.

\textbf{Politeness} Politeness means the degree of respectfulness and courtesy~\cite{song2025assessing}.
Politeness in responses is essential in communication, which helps to foster user engagement~\cite{shan2022language}.
A polite counterspeech helps avoid potential backlash and is more likely to be accepted by users, as it encourages a respectful tone~\cite{song2025echoes, yue2024evidence,he2023reinforcement}.
We utilize the Multilingual Politeness Classification Model, which is a computational tool designed to assess the politeness of responses~ \cite{srinivasan2022tydip}.


\textbf{Factual Accuracy} It measures the reliability and trustworthiness of the generated response by ensuring that the information provided is correct~\cite{zhou2024trustworthiness}. 
Presenting scientifically accurate facts in counterspeech can effectively correct misinformation and maintain user trust \cite{yue2024evidence}.
We employ LLM-based evaluation to assess whether the facts presented in counterspeech are correct.
While LLMs exist hallucinations, studies find that when appropriately prompted, they could be strong fact-checkers \cite{wang2025accuracy,guan2024language, chen2023evaluating}. Considering both the cost and need for updated information, we prompt gpt-4o-mini-2024-07-18\footnote{https://platform.openai.com/docs/models/gpt-4o-mini} with guidance, such as encouraging web search and explanation generation to assist with fact checking (Prompts detailed in \ref{Appendix: Factual Accuracy Prompt}).
To validate the reliability of the evaluation, we conduct a human assessment, which achieves a tolerant match score above 0.80 and a Cohen’s Kappa above 0.70, indicating strong reliability (See further details in Appendix \ref{Appendix: Evaluation Validation}).



\subsection{Results}
We present the evaluation of results in Table \ref{tab:Main_results}. There are several findings we conclude as follows.


\textbf{RAG helps improve the readability and factual accuracy, but shows limited gains in user preference.} RAG incorporates customized evidence into generation, which enhances the readability of responses (e.g., LLaMA-8B: RAG vs. Instructional Prompt: 1.94 vs. 2.74; Qwen-7B: 1.23 vs. 1.66 ), and factual accuracy (e.g., LLaMA-8B: RAG vs. Instructional Prompt: 0.89 vs. 0.97; Qwen-7B: 0.87 vs. 0.83;
LLaMA-1B: 0.66 vs. 0.54).
However, its impact on user preference is limited (LLaMA-8B: 0.71 vs. 0.75; Qwen-7B: 0.70 vs. 0.73; LLaMA-1B: 0.34 vs. 0.64).
It indicates that the customized evidence may help improve the readability at the syllabus and word level and provide more accurate facts, but the current integration may have limited capability to improve the generation quality, highlighting the need to further optimize the generation module.

\textbf{Controlled-Literacy achieves higher overall performance.}
Our Controlled-Literacy framework consistently achieves strong performance across all three evaluation dimensions. For LLaMA-8B, they demonstrate higher politeness (0.84), better alignment with target readability levels (Target Distance: 0.90), strong user preference (0.74), and higher factual accuracy (0.91).
Although Instructional Prompt slightly outperforms in user preference (0.75), it shows inferior performance in politeness (0.41), target alignment (2.74), and factual accuracy (0.87).
In the case of Qwen-7B and  LLaMA-1B, Controlled-Literacy achieves the best overall performance across all metrics: politeness (0.73 and 0.74), target distance (2.06 and 1.31), user preference (0.74 and 0.69), and factual accuracy (0.90 and 0.71).
Moreover, we observe that smaller models (e.g., LLaMA-1B) benefit more significantly from Controlled-Literacy, exhibiting greater improvements in overall performance.

\section{Cross Evaluation}
\label{main: Cross Generation}
We employ two evaluation methods to assess our proposed framework.
Firstly, we test our methods on \textit{Check-COVID} and \textit{MisinfoCorrect} to examine their generalization ability. These two datasets reflect different sources of misinformation: Twitter and online news media, respectively. 
Second, we perform cross-evaluation to analyze how users with a particular health literacy level respond to counterspeech generated for other literacy levels. This allows us to demonstrate that our method is more effective when tailored to the intended target health literacy level.

\begin{table}[ht]
\centering
\small
\resizebox{0.45\textwidth}{!}{
\begin{tabular}{lccc}
\toprule
\textbf{Counterspeech/User} & \textbf{Low User} & \textbf{Medium User} & \textbf{High User} \\
\midrule
\multicolumn{4}{l}{\textbf{Instructional Prompt }} \\
\quad low    & \cellcolor{blue!10} \textbf{0.75 (0.03)}     & \textbf{0.75 (0.04)}     & 0.68 (0.12) \\
\quad medium & 0.73 (0.08)    & \cellcolor{blue!10}0.74 (0.06)      & 0.74 (0.07) \\
\quad high   & 0.55 (0.16)    & 0.62 (0.15)     & \cellcolor{blue!10}\textbf{0.75 (0.04)} \\
\midrule
\multicolumn{4}{l}{\textbf{RAG}} \\
\quad low    & \cellcolor{yellow!30}\textbf{0.73 (0.09)}     & \textbf{0.73 (0.10)}     & 0.50 (0.17) \\
\quad medium & 0.72 (0.11)    & \cellcolor{yellow!30}\textbf{0.73 (0.11) }     & 0.66 (0.17) \\
\quad high   & 0.45 (0.18)    & 0.51 (0.18)     & \cellcolor{yellow!30}\textbf{0.72 (0.09)} \\
\midrule
\multicolumn{4}{l}{\textbf{\textit{Controlled-Literacy}}} \\
\quad low    &\cellcolor{green!20} \textbf{0.74 (0.07)}     & \textbf{0.74 (0.05)}     & 0.61 (0.16) \\
\quad medium & 0.69 (0.16)    & \cellcolor{green!20}0.73 (0.10)      & 0.62 (0.23) \\
\quad high   & 0.62 (0.15)    & 0.63 (0.13)     & \cellcolor{green!20}\textbf{0.75 (0.00)} \\
\bottomrule
\end{tabular}}
\caption{Cross evaluation of user preference.}
\label{tab:Cross evaluation}
\end{table}

\textbf{Controlled-Literacy generalizes robustly across datasets.} We report the results of the first evaluation in Table \ref{tab:checkcovid_results} and Table \ref{tab:misinfo_correct_results} in Appendix \ref{Appendix: Cross Generalization Results}. Across the CheckCovid and MisInfoCorrect datasets, the Controlled-Literacy method consistently achieves the best average performance in all four metrics: (1) Politeness: Highest across nearly all model configurations (e.g., 0.85 in LLaMA-8B for CheckCovid, 0.86 in LLaMA-8B for MisInfoCorrect). (2) Target Distance: Consistently the lowest among all baselines, indicating superior alignment with the intended readability level. (3) User Preference: Typically equal to or higher than other methods (e.g., 0.75 in most configurations). (4) Factual accuracy: Always higher than RAG and Instructional Prompt, which indicates that counterspeech generated by Controlled-Literacy is more reliable. 

The improvements in politeness, target distance, and factual accuracy are particularly notable in smaller models (e.g., LLaMA-1B), indicating that literacy control is especially beneficial for low-capacity settings when dealing with informal health misinformation.

\textbf{User preference peaks when counterspeech matches literacy level.} The second evaluation results in Table~\ref{tab:Cross evaluation} show that the top preference scores for each user group generally occur when the counterspeech matches the user's literacy level. For instance, in the Instructional Prompt setting, low and high users show the highest preference for low-level (0.75) and high-level (0.75) responses, respectively. Similarly, under the RAG setting, the top scores appear for low (0.73), medium (0.73), and high (0.75) users when aligned with their corresponding levels. The Controlled-Literacy method shows a similar trend, with low users preferring low-level counterspeech (0.74) and high users favoring high-level responses (0.75).

However, there are notable exceptions where users prefer counterspeech designed for different literacy levels. For example, in the Instructional Prompt setting, medium users rated low-level counterspeech slightly higher (0.75) than medium-level (0.74). In the RAG setting, medium users rated both low and medium-level counterspeech equally (0.73). Likewise, in the Controlled-Literacy setting, medium users showed a higher preference for low-level responses (0.74) compared to medium-level (0.73).
These findings suggest that while aligning counterspeech with user literacy level generally yields the best user preference outcomes, users may sometimes prefer responses with a lower readability level, possibly due to better accessibility. This inspires us to further optimize our generation strategy by aligning responses more closely with the lower end of the target health literacy range in the future.


\section{Qualitative Analysis}
To understand why the optimized counterspeech achieves higher performance across politeness, target distance, and factual accuracy, yet still falls short of a perfect user preference score, 
we conduct a qualitative analysis with human experts. We engage two experts in misinformation studies to review and analyze the reasons, identify their strengths, and highlight remaining limitations. We randomly select 50 samples from the best-performing model, Controlled-Literacy, using LLaMA3.1-8B-Instruct, for high health literacy level of users.

They summarize several key elements that make the counterspeech achieve high performance:
(1) Evidence-based reasoning: most of the counterspeech cite scientific sources (CDC, WHO, peer-reviewed studies) and explain the details, which matches the expectations of rational justification for high health literacy level of users. 
(2) Precise terminology. The counterspeech uses domain-specific vocabulary, such as ``pathogenesis,'' ``clinical trials'' and so on. 
(3) Structured argumentation. The counterspeech often follows a consistent structure: acknowledgement → clarification → evidence → implication → recommendation, satisfying expectations for logical coherence and depth.

However, our best-performing responses also exhibit several shortcomings that need further improvement.
(1) Repetitive openings. Many counterspeech begin with similar phrases (e.g., ``Thank you for sharing...''), which lack lexical variety. It tends to be less human-like, potentially decreasing the effectiveness of counterspeech.
(2) Heavy information density. While users with high health literacy generally prefer more technical content, some responses are overly dense and packed with information. It may diminish clarity and fatigue readers.
(3) Detached tone. Although the counterspeech maintains a coherent and logically structured academic tone, it often lacks emotional resonance or personal storytelling. Incorporating more empathetic or narrative elements could enhance reader engagement and strengthen the persuasive impact.

\section{Conclusion}
We propose a Controlled-Literacy method to create counterspeech that matches users' health literacy levels when addressing health misinformation.
This framework enables the integration of RAG and RL to generate accessible and audience-appropriate counterspeech. To control the evidence used during generation, we construct a knowledge base that contains diverse evidence and filter the evidence for each target group after retrieval.
Our reward design combines user preference and readability, ensuring the generated content aligns with the health literacy needs of different user groups. Experimental results show that Controlled Literacy produces counterspeech that is more accessible, polite, user-preferred by intended groups, and factually accurate. Furthermore, cross-generalization experiments demonstrate the robustness of our method across various types of health misinformation.

\section*{Limitations}
\textbf{Limited coverage in retrieved knowledge.} Although we collect knowledge from diverse sources, our current approach does not incorporate real-time information, which is critical for addressing dynamic and evolving health misinformation. In the future, we will incorporate web search into our generation to enhance the timeliness and relevance of retrieved knowledge.\\

\noindent
\textbf{Need for finer-grained user group categorization.} In our study, we only consider three health literacy level. However, real-world users exhibit a wide range of characteristics and information needs. We aim to develop a more nuanced and detailed user segmentation strategy to better align with diverse health literacy profiles.\\

\noindent
\textbf{Insufficient evaluation framework.} Our evaluation mostly relies on computing-based measurement. While informative, these do not fully capture how real users perceive or benefit from the generated counterspeech.  Moving forward, we plan to incorporate more human evaluations from diverse user groups to obtain  a comprehensive assessment of counterspeech effectiveness.

\section*{Ethics Statement}
We ensure that our study adheres to ethical guidelines by carefully evaluating associated risks and benefits. 
We collect data from Reddit under Reddit’s Terms of Service using PRW API. 
Reddit is a public forum. When users sign up to Reddit, they consent to make their data available to the third party, including the academy. Therefore, we can use Reddit data without further seeking user consent following the ethical rules \cite{procter2019study}. 
We have masked users' identifiable information before analysis and modeling.
We will make sure the dataset is exclusively used for non-commercial research purposes\footnote{\url{https://www.reddit.com/wiki/api-terms/}}. 
We acknowledge the potential risks of users being re-identified with anonymized data or misuse of the data by individuals, but the benefits will outweigh such risks.


\bibliography{custom}

\appendix

\section{Subreddits List}
\label{Appendix:SubredditsList}


Table~\ref{tab:subreddits} presents the list of subreddits analyzed in this study. These subreddits were selected to cover a broad spectrum of discussions on science, public health, medical topics, vaccines, and conspiracy theories.

\begin{table}[h!]
\begin{center}
\resizebox{\columnwidth}{!}{%
\begin{tabular}{l}
\toprule
\textbf{Subreddit Name} \\
\midrule
\texttt{r/science}, \texttt{r/Wuhan\_Flu}, \texttt{r/CoronavirusCirclejerk},\\ \texttt{r/DebateVaccines}, \texttt{r/vaccinelonghaulers}, \\
\texttt{r/ChurchofCOVID}, \texttt{r/vaccineautismevidence},\\ \texttt{r/ScienceUncensored}, \texttt{r/CoronavirusUS}, \texttt{r/conspiracy}, \\
\texttt{r/medical\_advice}, \texttt{r/joerogan}, \texttt{r/Conservative},\\ \texttt{r/vaccineaddiction}, \texttt{r/thingsprovaxxerssay} \\
\bottomrule
\end{tabular}%
}
\end{center}
\caption{Subreddit list
}
\label{tab:subreddits}
\end{table}





\section{Guidelines and Annotation process for detecting health misinformation}
\label{Appendix: detecting health misinformation}

\subsection{Annotation guidelines and process}
After collecting the Reddit dataset, the next step is to annotate posts into two categories: "Health Misinformation" and "Not Health Misinformation.". To ensure annotation consistency, all annotators received comprehensive guidelines, including examples of health misinformation, classification criteria, and citations of reputable sources (e.g., CDC, WHO, and NIH) for verification purposes.

\noindent\textbf{Annotation guidelines for health misinformation labeling}

The goal of this annotation task is to classify posts based on whether they 
contain health-related misinformation. Posts will be assigned one of two labels:

Label 1: "Health Misinformation", If the post contains health-related
misinformation.
Label 0: "Not Health Misinformation", If the post does not contain any 
health-related misinformation or is unrelated to health information.

Definition of Health Misinformation:
Any false, misleading, or unverified claim related to health, medicine, 
diseases, treatments, vaccines, nutrition, or wellness.
Misinformation includes claims that contradict established medical research,
public health guidelines, or authoritative sources such as the World Health
Organization (WHO) and the Centers for Disease Control and Prevention (CDC).

\subsection{Human validation}
\label{subsub:human_validation}
A human validation step was conducted to validate the reliability of these automatically labeled posts. Three annotators independently fact-checked a random sample of 100 model-labeled posts, assessing their accuracy. The agreement rate between the annotators is  88.1\%, 89.1\%, and 85.1\%, and Cohen’s Kappa scores are \( \kappa \geq 0.73 \), \( \kappa \geq 0.75 \), and \( \kappa \geq 0.67 \), respectively. This shows substantial agreement between the annotators. For the disagreement, we conducted further discussion and fact-checking, which concluded us with the final label. With the final label, our model agreement rate is 88.1\% and Cohen’s Kappa, \( \kappa \geq 0.73 \) demonstrates a substantial agreement between the model's classification and human judgment. 

\begin{table}[ht]
\centering
\resizebox{0.48\textwidth}{!}{
\begin{tabular}{ll}
\toprule
\textbf{Document Name} & \textbf{Source} \\
\midrule
COVID-19 Activity Book & Johns Hopkins Children's Center Child Life Program \\
Covid-19-Plain-Language-Flyer-with-Facemask & Independent Living Center of the Hudson Valley \\
COVID-19-Vaccines-A-Plain-Language-Guide & HealthMatters Program \\
FINAL Diagnostic Testing & U.S. Centers for Disease Control and Prevention \\
Healthy School Year & U.S. Centers for Disease Control and Prevention \\
Pediatric Testing Materials & Johns Hopkins Children's Center Child Life Program \\
Plain Language COVID Fact Sheets & Maryland Developmental Disabilities Council \\
Plain Language Guide for COVID-19\_Group-Home & St. Clair County Community Mental Health (SCCCMH) \\
COVID-19 Teen Info Sheet & U.S. Centers for Disease Control and Prevention \\
Symptoms Testing & U.S. Centers for Disease Control and Prevention \\
CDC Global Response to COVID-19\_CDC Archive & U.S. Centers for Disease Control and Prevention \\
CORD-19 & COVID-19 Open Research Dataset (CORD-19), Allen Institute for AI \\
COVID-19-Global-Response-fact-sheet & U.S. Centers for Disease Control and Prevention \\
FAQ COVID-19 Data and Surveillance\_CDC Archive & U.S. Centers for Disease Control and Prevention \\
\bottomrule
\end{tabular}}
\caption{Knowledge bases categorized by source}
\label{tab:knowledge-bases}
\end{table}

\section{Knowledge Bases}
\label{Appendix: Knowledge Bases}
We detail the knowledge bases designed for different health literacy levels in Table \ref{tab:knowledge-bases}.
We collect knowledge from a diverse set of reliable sources to ensure inclusiveness and representativeness across user groups, referring to previous studies~\cite{song2025dynamic,hong2025dynamic,anik2025multi}. These sources include federal health agencies such as the Centers for Disease Control and Prevention (CDC), academic and research institutions like the Johns Hopkins Children's Center and the Allen Institute for AI, as well as community organizations including the Maryland Developmental Disabilities Council, St. Clair County Community Mental Health (SCCCMH), and the Independent Living Center of the Hudson Valley. These organizations provide materials written in plain language, often tailored to individuals with diverse health literacy backgrounds.

\section{Prompt Design}
\label{appendix: Prompt Design}






\subsection{Instructional  Prompt}
\label{Appendix: Instructional  Prompt}

The Instructional  Prompt experiments aim to explore the full potential of LLMs in generating high-quality responses without relying on complementary techniques such as fine-tuning or external knowledge bases. To this end, we explicitly define the task and emphasize it at the beginning of each prompt to ensure clarity. 
We further design customized prompts tailored to different user groups, considering factors such as language style, evidence presentation, structural organization, and tone. This approach provides detailed and structured guidance to steer the LLMs toward generating responses that are both audience-appropriate and task-specific.

\paragraph{Low Health Literacy}
\begin{quote}

\texttt{<|Target Fkre|>80-100}\\
\texttt{<|Audience|> Low Health Literacy}\\
\texttt{<|Task|> Generate Counterspeech}

You are an expert in health communication and plain language.
Your audience has low health literacy — they have only basic reading and writing skills.

Your task is to write a counterspeech response to the following health misinformation, tailored specifically for this audience.

Your response must meet the following criteria:
\begin{enumerate}
    \item Simple and Clear Language: Use everyday words and short sentences. Avoid medical jargon and complex terms.
    \item Evidence-Based: Provide a fact-based correction in a way that’s easy to understand. 
    \item Clarity and Accessibility: Use simple examples or analogies to help explain your point.
    \item Polite and Respectful: Be kind and supportive. Do not shame or blame the person who may believe the misinformation. 
\end{enumerate}

Your response must be the counterspeech only — do not include any extra explanation or commentary.

Health misinformation to address: \texttt{"\{comment\}"}

\end{quote}

\begin{table*}[ht]
\centering
\small
\begin{tabular}{llcccc}
\toprule
\textbf{Health Literacy Level} & \textbf{\textit{Top-$k$}} & \textbf{Politeness} & \textbf{Target Distance} & \textbf{User Preference} & \textbf{Factual Accuracy} \\
\midrule
\multirow{3}{*}{Low} 
  & top\_10 & 0.72 (0.25) & 1.30 (3.23) & 0.71 (0.13) & \textbf{0.89} \\
  & top\_5  & \textbf{0.73 (0.25)} & \textbf{1.14 (2.97)} & 0.71 (0.12) & 0.87 \\
  & top\_3  & 0.71 (0.25) & 1.23 (3.14) & \textbf{0.73 (0.09)} & 0.84 \\
\hdashline
\multirow{3}{*}{Medium} 
  & top\_10 & 0.66 (0.22) & 4.93 (6.81) & 0.71 (0.13) & 0.81 \\
  & top\_5  & 0.70 (0.22) & 4.98 (6.76) & 0.72 (0.10) & 0.87 \\
  & top\_3  & \textbf{0.71 (0.23)} & \textbf{4.45 (6.35)} & \textbf{0.73 (0.11)} & \textbf{0.88} \\
\hdashline
\multirow{3}{*}{High} 
  & top\_10 & 0.41 (0.21) & 0.08 (0.75) & \textbf{0.72 (0.09)} & \textbf{0.90} \\
  & top\_5  & \textbf{0.43 (0.21)} & \textbf{0.04 (0.72)} & 0.71 (0.10) & 0.89 \\
  & top\_3  & 0.42 (0.21) & 0.42 (2.56) & 0.71 (0.10) & 0.81 \\
\bottomrule
\end{tabular}
\caption{Comparison of \textit{Top-$k$} selections.}
\label{tab:topk_eval}
\end{table*}

\paragraph{Medium Health Literacy}
\begin{quote}

\texttt{<|Target Fkre|>60-79}\\
\texttt{<|Audience|> Medium Health Literacy}\\
\texttt{<|Task|> Generate Counterspeech}

You are an expert in health communication with a focus on individuals with medium health literacy.  
This audience possesses the cognitive and social skills needed to actively participate in healthcare, communicate effectively with providers, and apply new information to changing circumstances
Your task is to generate a counterspeech response to a piece of health misinformation, tailored to this audience.

Your response should meet the following criteria:
\begin{enumerate}

\item Clear and Understandable Language: Use plain words and short, simple sentences. Avoid complex grammar. You may include basic medical terms, but explain them clearly and briefly. 
\item Evidence-Based Correction: Give a fact-based explanation using trusted health information. Keep the explanation short, logical, and easy to follow.  
\item Organized and Structured: Present your response in a simple and clear format. Use short paragraphs or bullet points if needed.  
\item Polite and Respectful: Be kind and supportive. Do not shame or blame the person who may believe the misinformation. 
\end{enumerate}
Your response must be the counterspeech only — do not include any extra explanation or commentary.

Health misinformation to address: \texttt{"\{comment\}"}

\end{quote}

\paragraph{High Health Literacy}
\begin{quote}

\texttt{<|Target Fkre|>0-59}\\
\texttt{<|Audience|> High Health Literacy}\\
\texttt{<|Task|> Generate Counterspeech}

You are an expert in health communication and digital literacy, specializing in engaging audiences with high health literacy who encompasses the ability to critically analyze information, understand social determinants of health, and engage in collective actions to address health disparities. 
Your task is to generate a counterspeech response to a piece of health misinformation.  

Your response should meet the following criteria:
\begin{enumerate}
\item  Advanced Language: Use precise, nuanced language that reflects the audience's ability to analyze, synthesize, and apply complex health information.
\item Evidence-Based Correction: Correct the misinformation with accurate, research-backed health information.
\item Clarity and Depth: Employ clear, well-structured arguments and sophisticated examples or analogies that resonate with an informed audience.
\item Polite and Respectful: Be kind and supportive. Do not shame or blame the person who may believe the misinformation.
\end{enumerate}
Your response must be the counterspeech only — do not include any extra explanation or commentary.

Health misinformation to address: \texttt{"\{comment\}"}

\end{quote}

\subsection{Evaluation Prompt - User Preference}
\label{Appendix: User Preference Prompt}
\paragraph{Low Health Literacy}
\begin{quote}
Assume you are a user with low health literacy—someone who may struggle to understand basic health information such as medication labels, appointment slips, or preventive care guidelines. You are presented with a piece of health misinformation and a counterspeech response written for your understanding.

\texttt{Misinformation\_Comment: "\{misinfo\_comment\}"}\\
\texttt{Counterspeech\_Response: "\{counterspeech\}"}

Evaluate the counterspeech based on its clarity and effectiveness in correcting the misinformation.

Use the following 1–5 Likert-style scale:
\begin{itemize}
  \item 1: Poor – Very difficult to understand and does little to correct the misinformation.
  \item 2: Fair – Somewhat understandable but leaves confusion or only partially corrects the misinformation.
  \item 3: Good – Generally clear and corrects the misinformation to a fair extent.
  \item 4: Very Good – Clear, easy to understand, and effectively corrects the misinformation.
  \item 5: Excellent – Extremely clear, very easy to understand, and completely corrects the misinformation.
\end{itemize}

Provide only the score (an integer from 1 to 5) as your final output.
\end{quote}

\paragraph{Medium Health Literacy }
\begin{quote}
Assume you are a user with medium health literacy—someone who can understand and act on straightforward health information but may struggle with complex or abstract concepts. You are presented with a piece of health misinformation and a counterspeech response written for your understanding.

\texttt{Misinformation\_Comment: "\{misinfo\_comment\}"}\\
\texttt{Counterspeech\_Response: "\{counterspeech\}"}

Evaluate the counterspeech based on its clarity and effectiveness in correcting the misinformation.

Use the following 1–5 Likert-style scale:
\begin{itemize}
  \item 1: Poor – Overly complex or ambiguous, difficult to understand and fails to correct the misinformation.
  \item 2: Fair – Somewhat clear but includes complexity that hinders full understanding.
  \item 3: Good – Generally clear and reasonably corrects the misinformation, though some parts may be slightly complex.
  \item 4: Very Good – Clear, straightforward, and effectively corrects the misinformation with minimal complexity.
  \item 5: Excellent – Extremely clear, easy to understand, and fully corrects the misinformation in an accessible way.
\end{itemize}

Provide only the score (an integer from 1 to 5) as your final output.
\end{quote}

\paragraph{High Health Literacy}
\begin{quote}
Assume you are a user with high health literacy—someone capable of analyzing, synthesizing, and applying complex health information across diverse contexts. You are presented with a piece of health misinformation and a counterspeech response written for your understanding.

\texttt{Misinformation\_Comment: "\{misinfo\_comment\}"}\\
\texttt{Counterspeech\_Response: "\{counterspeech\}"}

Evaluate the counterspeech based on its clarity and effectiveness in correcting the misinformation.

Use the following 1–5 Likert-style scale:
\begin{itemize}
  \item 1: Poor – Oversimplified or incomplete, lacking sufficient depth to correct the misinformation.
  \item 2: Fair – Addresses the misinformation but misses nuances or provides a partial correction.
  \item 3: Good – Generally clear and corrects the misinformation adequately, though some complexity may be missing.
  \item 4: Very Good – Clear, comprehensive, and effectively corrects the misinformation with a well-balanced explanation.
  \item 5: Excellent – Extremely clear, insightful, and provides a nuanced, well-supported correction that fully addresses the complexities.
\end{itemize}

Provide only the score (an integer from 1 to 5) as your final output.
\end{quote}

\subsection{Evaluation Prompt - Factual Accuracy}
\label{Appendix: Factual Accuracy Prompt}
\begin{quote}
You are an expert fact-checker. Your task is to determine whether the following counter-speech is factually correct.  
You may search the web to verify the claims made in the response.

\textbf{Counter-Speech Response:}  \\
\texttt{"\{model\_response\}"}

\textbf{Evaluation Instructions:}
\begin{itemize}
    \item If the counter-speech is factually correct and does not contain misinformation, output: \texttt{Label: 1}
    \item If the counter-speech contains false or misleading claims, output: \texttt{Label: 0}
\end{itemize}

Provide the label and explanations.

\textbf{Output Format:}  \\
\texttt{Label: (0 or 1)}\\
\texttt{Explanations: }
\end{quote}

\begin{table*}[htb!]
\centering
\small
\renewcommand{\arraystretch}{0.7}
\resizebox{\textwidth}{!}{
\begin{tabular}{llcccc}
\toprule
\textbf{Method} & \textbf{Literacy Level} & \textbf{Politeness} & \textbf{Target Distance (↓)} & \textbf{User Preference} & \textbf{Factual Accuracy}\\
\midrule

\multicolumn{5}{l}{\textbf{LLaMA-8B}} \\

Instructional Prompt  
    & low     & 0.52 (0.19) & 2.69 (4.25)  & 0.74 (0.06) &0.87\\
    & medium  & 0.60 (0.17) & 4.94 (7.49)  & 0.74 (0.08) &0.86\\
    & high    & 0.51 (0.12) & 0.03 (0.56)  & 0.74 (0.07) &0.76\\
    & \textbf{Avg.} & \textbf{0.54 (0.16)} & \textbf{2.55(4.10)} & \textbf{0.74 (0.07)} & \textbf{0.83}\\


RAG       
    & low     & 0.72 (0.19) & 2.11 (4.06)  & 0.74 (0.06) &0.82\\
    & medium  & 0.74 (0.16) & 3.21 (5.82)  & 0.74 (0.07) &0.85\\
    & high    & 0.57 (0.13) & 0.05 (0.60)  & 0.70 (0.11) &0.88\\
    & \textbf{Avg.} & \textbf{0.68 (0.16)} & \textbf{1.79 (3.49)} & \textbf{0.73 (0.08)} & \textbf{0.85}\\

\textbf{\textit{Controlled-Literacy}}

    & low     & 0.82 (0.12) & 0.74 (1.90)  & 0.75 (0.00) &0.92\\
    & medium  & 0.75 (0.15) & 2.40 (4.26)  & 0.75 (0.02) &0.91\\
    & high    & 0.99 (0.00) & 0.00 (0.00)  & 0.75 (0.02) &0.96\\
    & \cellcolor{gray!20}\textbf{Avg.} & \cellcolor{gray!20}\textbf{0.85 (0.09)} & \cellcolor{gray!20}\textbf{1.05 (2.05)} & \cellcolor{gray!20}\textbf{0.75 (0.01)} & \cellcolor{gray!20}\textbf{0.93}\\

\midrule
\multicolumn{5}{l}{\textbf{LLaMA-1B}} \\

Instructional Prompt   
    & low     & 0.68 (0.21) & 4.36 (7.40)  & 0.65 (0.19) &0.57\\
    & medium  & 0.50 (0.19) & 17.29 (11.74) & 0.55 (0.23) &0.59\\
    & high    & 0.84 (0.17) & 0.06 (0.60)  & 0.65 (0.16) &0.46\\
    & \textbf{Avg.} & \textbf{0.67 (0.19)} & \textbf{7.24 (6.58)} & \textbf{0.62 (0.19)} &\textbf{0.54}\\


RAG         
    & low     & 0.52 (0.19) & 16.90 (14.68) & 0.41 (0.24) &0.69\\
    & medium  & 0.55 (0.15) & 9.19 (10.50)  & 0.43 (0.23) &0.59\\
    & high    & 0.57 (0.17) & 0.11 (1.27) & 0.43 (0.21) &0.51\\
    & \textbf{Avg.} & \textbf{0.55 (0.17)} & \textbf{8.73 (8.82)} & \textbf{0.42 (0.23)} &\textbf{0.60}\\


\textbf{\textit{Controlled-Literacy}}
    & low     & 0.80 (0.14) & 2.29 (3.69)  & 0.71 (0.13) &0.69\\
    & medium  & 0.79 (0.17) & 2.22 (3.81)  & 0.72 (0.10) &0.74\\
    & high    & 0.83 (0.27) & 0.05 (0.73)  & 0.68 (0.13) &0.75\\
    & \cellcolor{gray!20}\textbf{Avg.} & \cellcolor{gray!20}\textbf{0.81 (0.19)} & \cellcolor{gray!20}\textbf{1.52 (2.74)} & \cellcolor{gray!20}\textbf{0.70 (0.12)} &\cellcolor{gray!20}\textbf{0.73}\\

\midrule
\multicolumn{5}{l}{\textbf{Qwen-7B}} \\

Instructional Prompt  
    & low     & 0.46 (0.18) & 4.93 (7.12)  & 0.73 (0.10) &0.84\\
    & medium  & 0.59 (0.18) & 4.79 (7.69)  & 0.74 (0.11) &0.95\\
    & high    & 0.53 (0.13) & 0.58 (6.05)  & 0.74 (0.10) &0.97\\
    & \textbf{Avg.} & \textbf{0.53 (0.16)} & \textbf{3.43 (6.95)} & \textbf{0.74 (0.10)} &\textbf{0.92}\\


RAG         
    & low     & 0.58 (0.17) & 5.20 (6.98)  & 0.74 (0.07) &0.88\\
    & medium  & 0.64 (0.16) & 7.09 (8.22)  & 0.75 (0.04) &0.95\\
    & high    & 0.63 (0.14) & 0.03 (0.38)  & 0.73 (0.08) &0.92\\
    & \textbf{Avg.} & \textbf{0.62 (0.16)} & \textbf{4.11 (5.19)} & \textbf{0.74 (0.06)} &\textbf{0.92}\\

\textbf{\textit{Controlled-Literacy}}
    & low     & 0.79 (0.16) & 1.94 (3.71)  & 0.75 (0.04) &0.90\\
    & medium  & 0.62 (0.13) & 1.36 (2.67)  & 0.75 (0.04) &0.97\\
    & high    & 0.88 (0.18) & 0.00 (0.00)  & 0.74 (0.05) &0.98\\
    & \cellcolor{gray!20}\textbf{Avg.} & \cellcolor{gray!20}\textbf{0.76 (0.16)} & \cellcolor{gray!20}\textbf{1.10 (2.13)} & \cellcolor{gray!20}\textbf{0.75 (0.04)} &\cellcolor{gray!20}\textbf{0.95}\\

\bottomrule
\end{tabular}}
\caption{Cross generalization performance on \textbf{CheckCovid Dataset}. The best overall performance in each category is highlighted in gray.}
\label{tab:checkcovid_results}
\end{table*}

\section{Evaluation Validation}
\label{Appendix: Evaluation Validation}
Considering that \textbf{user preference} evaluations rely on LLMs, which may introduce potential bias and may not fully capture real users' perspectives, we additionally conduct a human evaluation. We recruit six annotators representing low, medium, and high health literacy levels.
To select annotators with diverse health literacy, we administer a brief screening survey following the method proposed by \citet{rasmussen2023brief}.
Using this method, respondents rate each item on a 4-point scale using the following categories: Not at all true (1 point), Not completely true (2 points), Somewhat true (3 points), and Absolutely true (4 points). A cumulative score is then calculated. Based on the scoring criteria from the HLSAC, the initial version of the B-HLA categorizes health literacy as follows: low health literacy (10–25 points), moderate health literacy (26–35 points), and high health literacy (36–40 points).
\begin{table*}[htb!]
\centering
\small
\renewcommand{\arraystretch}{0.7}
\resizebox{\textwidth}{!}{
\begin{tabular}{llccccc}
\toprule
\textbf{Method} & \textbf{Literacy Level} & \textbf{Politeness} & \textbf{Target Distance (↓)} & \textbf{User Preference}& \textbf{Factual Accuracy} \\
\midrule
\multicolumn{5}{l}{\textbf{LLaMA-8B}} \\

Instructional Prompt    
& low     & 0.41 (0.24) & 2.17 (4.03)  & 0.75 (0.02) &0.98\\
& medium  & 0.42 (0.20) & 2.94 (5.59)  & 0.75 (0.01) &0.98\\
& high    & 0.29 (0.17) & 0.11 (0.93)  & 0.75 (0.04) &0.99\\
\textbf{} & \textbf{Avg.} 
& \textbf{0.37 (0.20)} 
& \textbf{1.74 (3.52)} 
& \textbf{0.75 (0.02)} 
& \textbf{0.98}\\

RAG              

& low     & 0.86 (0.15) & 0.98 (2.55)  & 0.75 (0.00) &0.99\\
& medium  & 0.72 (0.24) & 1.49 (2.84)  & 0.75 (0.03) &1.00\\
& high    & 0.34 (0.18) & 0.13 (1.07)  & 0.74 (0.04) &0.97\\
\textbf{} & \textbf{Avg.} 
& \textbf{0.64 (0.19)} 
& \textbf{0.87 (2.15)} 
& \textbf{0.75 (0.02)} 
& \textbf{0.99}\\

\textbf{\textit{Controlled-Literacy}} 
& low     & 0.92 (0.09) & 0.78 (1.95)  & 0.75 (0.00) &1.00\\
& medium  & 0.68 (0.23) & 0.89 (2.45)  & 0.75 (0.00) &1.00\\
& high    & 0.97 (0.02) & 0.00 (0.00)  & 0.75 (0.00) &1.00\\
\textbf{} & \cellcolor{gray!20}\textbf{Avg.} 
& \cellcolor{gray!20}\textbf{0.86 (0.11)} 
& \cellcolor{gray!20}\textbf{0.56 (1.47)} 
& \cellcolor{gray!20}\textbf{0.75 (0.00)} 
& \cellcolor{gray!20}\textbf{1.00}\\

\midrule

\multicolumn{5}{l}{\textbf{LLaMA-1B}} \\

Instructional Prompt    
& low     & 0.73 (0.28) & 12.41 (12.03) & 0.64 (0.18) &0.62\\
& medium  & 0.47 (0.32) & 15.82 (16.05) &0.59 (0.22) &0.72\\
& high    & 0.82 (0.18) & 0.14 (2.54)  & 0.68 (0.15) &0.78\\
\textbf{} & \textbf{Avg.} & \textbf{0.67 (0.26)} & \textbf{9.46 (10.21)} & \textbf{0.64 (0.18)} & \textbf{0.71}\\

RAG                  
& low     & 0.43 (0.30) & 15.53 (12.60) & 0.53 (0.23) &0.68\\
& medium  & 0.36 (0.25) & 5.59 (7.89)  & 0.48 (0.23) &0.79\\
& high    & 0.33 (0.23) & 0.51 (3.78)  & 0.49 (0.21) &0.76\\
\textbf{} & \textbf{Avg.} & \textbf{0.37 (0.26)} & \textbf{7.21 (8.09)} & \textbf{0.41 (0.22)} & \textbf{0.74}\\

\textbf{\textit{Controlled-Literacy}} 
& low     & 0.84 (0.16) & 4.54 (5.18)  & 0.68 (0.17) &0.74\\
& medium  & 0.61 (0.23) & 1.05 (2.30)  & 0.72 (0.08) &0.70\\
& high    & 0.92 (0.20) & 0.00 (0.00)  & 0.68 (0.13) &0.81\\
\textbf{} & \cellcolor{gray!20}\textbf{Avg.} & \cellcolor{gray!20}\textbf{0.79 (0.20)} & \cellcolor{gray!20}\textbf{1.86 (2.49)} & \cellcolor{gray!20}\textbf{0.69 (0.13)} & \cellcolor{gray!20}\textbf{0.75}\\

\midrule
\multicolumn{5}{l}{\textbf{Qwen-7B}} \\

Instructional Prompt    
& low     & 0.31 (0.22) & 3.04 (5.74)  & 0.74 (0.06) &0.91\\
& medium  & 0.45 (0.23) & 4.74 (10.06)  & 0.74 (0.10) &0.97\\
& high    & 0.38 (0.21) & 0.02 (0.41)  & 0.74 (0.05) &1.00\\
& \textbf{Avg.} & \textbf{0.38 (0.22)} & \textbf{2.60 (5.40)} & \textbf{0.74 (0.07)} &\textbf{0.96}

\\

RAG                  
& low     & 0.59 (0.22) & 3.96 (6.38)  & 0.75 (0.01) &0.97\\
& medium  & 0.52 (0.22) & 4.84 (7.15)  & 0.75 (0.03) &0.96\\
& high    & 0.49 (0.26) & 0.02 (0.17)  & 0.74 (0.04) &1.00\\
& \textbf{Avg.} & \textbf{0.53 (0.23)} & \textbf{2.94 (4.57)} & \textbf{0.75 (0.03)} &\textbf{0.98}\\

\textbf{\textit{Controlled-Literacy}} 
& low     & 0.78 (0.19) & 2.32 (2.99)  & 0.75 (0.00) &1.00\\
& medium  & 0.43 (0.16) & 1.23 (2.53)  & 0.75 (0.03) &0.99\\
& high    & 0.77 (0.30) & 0.00 (0.00)  & 0.75 (0.00) &1.00\\
& \cellcolor{gray!20}\textbf{Avg.} & \cellcolor{gray!20}\textbf{0.66 (0.22)} & \cellcolor{gray!20}\textbf{1.18 (1.84)} & \cellcolor{gray!20}\textbf{0.75 (0.01)} &\cellcolor{gray!20}\textbf{1.00}
\\
\bottomrule
\end{tabular}}
\caption{Cross generalization performance on \textbf{MisinfoCorrect  Dataset}. The best overall performance in each category is highlighted in gray.}
\label{tab:misinfo_correct_results}
\end{table*}

Two annotators are selected from each health literacy category, resulting in a balanced sample across literacy levels. All annotators are proficient in English and have no prior involvement in the project.
Annotators are compensated at a rate of \$10 per hour, in accordance with ethical guidelines for human-subject research. Each annotator completes the evaluation in approximately two hours.
The final annotator group consisted of individuals aged 15 to 35 years, with educational backgrounds ranging from high school to graduate-level studies. The group includes three female and three male participants, all based in the United States. All annotators reported regular access to online health information, though their confidence and comprehension varied, as reflected in the screening tool scores.

We randomly sample 50 health misinformation paired with low/medium/high counterspeech generated by Instructional Prompt using LLaMA3.2-1B-Instruct. The evaluation guidelines provided to the LLM are also given to the human annotators to ensure consistency. 
To assess inter-annotator agreement, we compute both a tolerant match rate, acknowledging the difficulty of achieving exact agreement on a 1–5 scale, and the weighted Cohen’s Kappa, which accounts for the degree of disagreement by penalizing larger rating discrepancies more heavily.

Additionally, we recruit two annotators to assess the \textbf{factual accuracy} of the generated counterspeech.
Annotators are instructed as follows:
You are an expert fact-checker. Your task is to evaluate whether the following counter-speech is factually correct.
You may search the web to verify the claims made in the counter-speech.

\textbf{Evaluation Instructions:}

* If the counterspeech is factually correct and does not contain misinformation, label it as 1.

* If the counterspeech contains false or misleading claims, label it as 0.

You may consult fact-checking sources such as Snopes, HealthFeedback, and FactCheck.org to support your judgment.
Please provide a brief explanation for each label you assign.

\textbf{Correlation Analysis}
 We conducted a correlation analysis between human ratings and LLM-generated ratings of user preference using Pearson~\cite{benesty2009pearson}, Spearman, and Kendall’s Tau~\cite{kendall1938new}, referring to \citet{shen2023large}. 

We present the results in Table \ref{tab:correlation}.
The results show that Pearson correlation is highest for low and medium literacy levels (0.67–0.68), suggesting the LLM matches human scores most closely in absolute value for users with lower literacy.
Spearman and Kendall's Tau are highest at the high literacy level, indicating that the LLM is especially good at ranking outputs in the same order as human annotators at this level, even if the exact scores differ.
These findings suggest that the LLM’s behavior adapts well to different user profiles:
(1) For low and medium literacy users, the focus is on numeric simplicity and readability, and LLM-generated scores align well with human ratings in both value and order. (2) For high literacy users, while the Pearson correlation is slightly lower, the LLM captures the ranking preferences of more sophisticated users very effectively.

\section{\textit{Top-$k$} Comparison}
\label{Appendix: Top-k Comparison}
We use the LLaMA3.1-8B-Instruct model within a RAG framework to examine the impact of Top-$k$ evidence selection. Results in Table~\ref{tab:topk_eval} reveal the following: for users with low health literacy, \textit{Top-5} achieves the highest politeness score (0.73) and lowest target distance (1.14), \textit{Top-3} yields the highest user preference score (0.73), and \textit{Top-10} leads in factual accuracy (0.89).
Given the relatively small differences in politeness, user preference, and target distance but a more substantial advantage in factual accuracy, we select \textit{Top-10} for users with low health literacy.
For users with medium health literacy, \textit{Top-3} achieves the best overall performance across all dimensions: politeness (0.71), target distance (4.45), user preference (0.73), and factual accuracy (0.88), and is therefore selected.
In the high health literacy setting, \textit{Top-10} is chosen as it demonstrates the best performance in user preference (0.72) and factual accuracy (0.90).

\section{Cross Generalization Results}
\label{Appendix: Cross Generalization Results}
We apply our methods, Controlled-Literacy, to two distinct misinformation datasets: Check-COVID and MisinfoCorrect mentioned in Section \ref{main: Cross Generation}. The results are in Table \ref{tab:checkcovid_results} and Table \ref{tab:misinfo_correct_results}.

\section{Computing Resources}
The computational resources applied in this research include a high-performance server equipped with an Intel Xeon Gold 6226R processor, 128 GB memory, and 3 Nvidia RTX 8000 GPUs.

\section{Use of AI Assistants}
We acknowledge the use of AI tools to assist with code writing and expression refinement. 
The authors developed all core ideas, methods, analyses, and conclusions. 
The final content reflects the authors' independent scholarly contributions.

\end{document}